\documentclass[preprint,12pt]{elsarticle} 

\usepackage{amssymb}

\makeatletter
\newcommand{\removelatexerror}{\let\@latex@error\@gobble}
\makeatother

\usepackage{hyperref}
\usepackage{url}
\usepackage{subcaption}
\usepackage{amssymb}
\usepackage{amsmath}
\usepackage{amsthm,amsmath,amssymb}
\usepackage{mathrsfs}

\usepackage[flushleft]{threeparttable}
\usepackage{threeparttable}
\usepackage{booktabs}
\usepackage{comment}
\usepackage{amsthm}
\usepackage{amsmath}
\usepackage{apxproof}
\usepackage{thmtools, thm-restate}
\usepackage{wrapfig}
\usepackage{adjustbox}
\usepackage[flushleft]{threeparttable}
\usepackage{graphicx}

\usepackage{dashrule}

\usepackage[noamssymbols,upint]{newtxmath} 

\usepackage[ruled,linesnumbered]{algorithm2e}
\usepackage{algorithmic}
\usepackage{float}
\usepackage{ulem}
\usepackage{cancel}

\DeclareMathAlphabet\mathbfcal{OMS}{cmsy}{b}{n}

\usepackage{natbib}

\usepackage{microtype}
\usepackage{graphicx}

\usepackage{booktabs} 

\usepackage{hyperref}
\usepackage{amssymb}
\usepackage{amsmath}
\usepackage{amsthm}

\usepackage{adjustbox}

\usepackage{cancel}

\usepackage[inline]{enumitem}

\usepackage{bm}
\usepackage{makecell}
\usepackage[figuresright]{rotating}

\usepackage{adjustbox}

\usepackage{enumitem}

\usepackage{graphicx}
\usepackage{array}

\usepackage{extarrows}
\usepackage{caption}
\usepackage{graphicx}
\usepackage{sidecap}
\usepackage{float}
\usepackage{framed}
\usepackage{tabularx}
\usepackage{twoopt}
\usepackage{adjustbox}

\usepackage{lscape}

\usepackage[table]{xcolor}

\usepackage{tikz}
\usetikzlibrary{fit}

\usepackage{booktabs,threeparttable}

\usepackage{lineno}

\newcommandtwoopt\Textbox[5][7.2cm][2cm]{%
\begin{tikzpicture}[remember picture,overlay]
  \coordinate (aux) at ([xshift=#1]#4);
  \node[inner ysep=3pt,yshift=1ex,draw=pink,thick,
    fit=(#3) (aux),baseline] 
    (box) {};
  \node[text width=#2,anchor=north east,
    font=\sffamily\footnotesize,
  align=right
    ] 
    at (box.north east) {#5};
\end{tikzpicture}%
}

\hypersetup{
colorlinks=true,
citecolor=red,
    linkcolor=red,
    filecolor=magenta,      
    urlcolor=red,
linktocpage}

\usepackage{lipsum}


\begin{document}

\begin{frontmatter}

\title{Mutual Enhancement of Large Language and Reinforcement Learning Models through Bi-Directional Feedback Mechanisms: A Planning Case Study}





\author{Shangding Gu$^{*}$
}
\cortext[cor1]{We appreciate any constructive
comments and suggestions corresponding to \textit{gushangding@gmail.com}. } 


\begin{abstract}
Large Language Models (LLMs) have demonstrated remarkable capabilities for reinforcement learning (RL) models, such as planning and reasoning capabilities. However, the problems of LLMs and RL model collaboration still need to be solved. In this study, we employ a teacher-student learning framework to tackle these problems, specifically by offering feedback for LLMs using RL models and providing high-level information for RL models with LLMs in a cooperative multi-agent setting. Within this framework, the LLM acts as a teacher, while the RL model acts as a student. The two agents cooperatively assist each other through a process of {\textit{recursive help}}, such as ``I help you help I help." The LLM agent supplies abstract information to the RL agent, enabling efficient exploration and policy improvement. In turn, the RL agent offers feedback to the LLM agent, providing valuable, real-time information that helps generate more useful tokens. This bi-directional feedback loop promotes optimization, exploration, and mutual improvement for both agents, enabling them to accomplish increasingly challenging tasks. Remarkably, we propose a practical algorithm to address the problem and conduct empirical experiments to evaluate the effectiveness of our method.
\end{abstract}




\end{frontmatter}




\section{Introduction}
\label{section:introduction}

Large Language Models (LLMs)~\citep{openai2023gpt4, chang2023survey} have shown exceptional performance across various domains. Notably, LLMs are useful for applications like robot planning~\citep{singh2023progprompt}, machine translation~\citep{zhang2023prompting} and medicine~\citep{thirunavukarasu2023large}. In parallel, RL has demonstrated remarkable capabilities in various domains, including achieving human-level performance in games such as the game of Go~\citep{silver2016mastering} and multi-player poker~\citep{brown2019superhuman}. LLMs have been increasingly incorporated to enhance the performance of RL \citep{du2023guiding, szot2023large}. Likewise, RL has also been employed to augment the capabilities of LLMs, furthering their effectiveness \citep{ouyang2022training, gu2024teamsrl}.
Nevertheless, the effective harnessing of LLMs' latent potential in solving complex tasks, through the synergistic integration with powerful RL frameworks~\citep{sutton2018reinforcement}, remains a formidable challenge. Therefore, a critical question that emerges in this field is: How does an RL model cooperate with LLMs to perform a given task effectively?

Facilitating cooperation between RL models and LLMs requires mutually beneficial actions, leading to enhancing each model's performance. However, it is challenging to meet these needs due to RL models' and LLMs' distinct decision-making characteristics. Specifically, the capabilities of LLMs generally exceed those of RL models, indicating the importance of developing methodologies for instructing RL models to acquire high-level knowledge and ensuring RL models can deliver real-time feedback to LLMs.

In this study, to address the above challenge, we propose a teacher-student learning framework in a cooperative game, where the integration of RL models (students) and LLMs (teachers) with bi-directional feedback~\citep{gu2023human} may be an effective solution. The two models cooperatively carry out complex tasks, which can be considered a win-win collaboration, where the RL model and the LLM act as two agents, cooperating to complement, assist, and provide feedback to each other, ultimately solving the problem together.

\section{Related Work}

The relation between RL models and language representation is investigated in several methods \citep{akyurek2023rl4f, chen2023introspective, gu2024teamsrl, jiang2019language, ouyang2022training, shinn2023reflexion, uc2023survey, yang2021safe, zhao2023babystories}. For instance, in the work of \cite{chen2023introspective}, they deploy LLMs as an optimization objective in an RL decision-making loop. Then, they further make LLMs reflect their decision results to refine LLMs' output using prompt engineering tips. In the work of \cite{yang2021safe}, they leverage text as safety constraints for RL safe exploration. In the work of \cite{ouyang2022training}, they train LLMs to align with human values in an RL process, where the reward model is a supervised label to teach LLMs to follow human instructions.

The research most related to our study includes the works of \cite{carta2023grounding} and \cite{tran2023exploring}. In the work of \cite{carta2023grounding}, they employ LLMs as RL policies to acquire task-solving capabilities while learning new knowledge through interactive experiences. Their experimental findings suggest that their method outperforms baseline approaches. However, a potential limitation of their work is the absence of instruction feedback from RL models, which may impact the overall effectiveness of their method. In the work of \cite{tran2023exploring}, they deploy RL to train a conversational agent using a simulator and an initial text generated by a generative chat agent. Subsequently, they input the data from the RL-trained agent to the generative chat agent. Although their experiment results demonstrate that their method performs better than baselines, a concern of this approach may be the potential time consumption associated with RL training for multi-turn conversations, as each conversation may necessitate RL training requests. Additionally, achieving self-online learning for task execution could be challenging in this framework.

\section{Method}
\label{section:method}

\begin{figure}[htbp!]
 \centering
 {
\includegraphics[width=0.39\linewidth, angle=0]{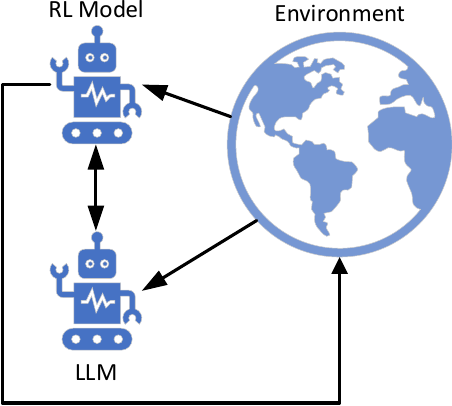}
}
 {
\includegraphics[width=0.55\linewidth]{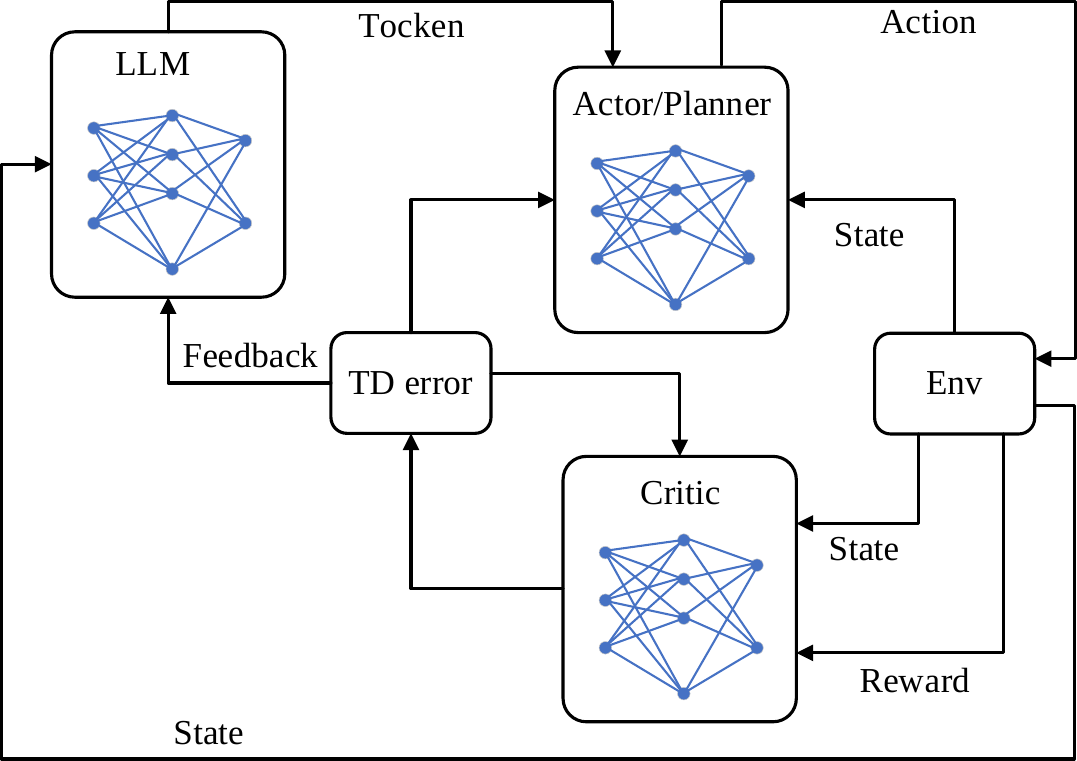}
}
    \vspace{-0pt}
 	\caption{\normalsize 
  An LLM and an RL model collaboratively engage with an environment to accomplish complex tasks, facilitating bi-directional feedback throughout the process.
 	} 
  \label{fig:LLM-RL-Model-Environments}
 \end{figure}

In this section, we introduce a teacher-student learning framework with bi-directional feedback, wherein a synergistic partnership between an LLM and an RL model is employed to tackle tasks collaboratively. As illustrated in Figure~\ref{fig:LLM-RL-Model-Environments}, these two models operate in tandem, with mutual support, ultimately enabling successful task completion~\footnote{We use a TD error as an estimator for the case's advantage function.}.

\textbf{LLMs (teachers) help RL models (students)}: LLMs often struggle to generate instructions that fully capture detailed and precise environmental information. However, they can still provide approximate guidance to RL models, aiding the exploration process. By narrowing the exploration space and accelerating policy discovery, such guidance improves the efficiency of RL training. This highlights the potential of LLMs to mitigate challenges arising from imperfect instructions, thereby enhancing RL performance in policy optimization.

\textbf{RL models (students) help LLMs (teachers)}: During policy execution in the RL framework, RL models benefit from the guidance provided by LLMs. In this collaborative process, RL models not only utilize but also evaluate the outputs generated by LLMs. This reciprocal interaction enables RL models to provide constructive feedback, facilitating the iterative refinement of LLM performance. As iterations progress, LLMs can better understand the environment, allowing them to generate increasingly effective guidance. This, in turn, enhances the ability of both RL models and LLMs to tackle complex tasks with greater efficiency. The iterative relationship between RL models and LLMs highlights their potential for continuous improvement and optimization. The corresponding algorithm is presented in Algorithm~\ref{algorithm:RL-LLM-algorithm-simple}.

\begin{algorithm}[!ht]
\caption{Onling Learning to make decisions with RL and LLMs.}
\label{algorithm:RL-LLM-algorithm-simple}
\begin{algorithmic}[1]
\STATE Initial Q value $Q_{\theta_0}$, V value $V_{\phi_0}$, and advantage function value $A_0^P = Q_{\theta_0} - V_{\phi_0}$, state $s_0$, action $a_0$, token $x_0$.
\FOR{$t=0, 1, \dots, T$} 
    \STATE Conduct tasks with an RL model $P_{\theta_t}(a_t \mid x_t, s_t)$ and an LLM $M(x_t \mid s_t)$.
    \STATE LLM $M(x_t \mid s_t)$ provides decision information to RL model $P_{\theta_t}(a_t \mid x_t, s_t)$ by leveraging comsense capabilities and environment information.
    \STATE Estimate new RL $A_t^P{'}$ based on new Q value $Q_{\theta_t}$ and V value $V_{\phi_t}$. 
    \IF{$A^P > A_t^P{'}$ }
    \STATE Provide negative instruction feedback to LLM, the token is worse than the last one.
    \ELSE 
    \STATE Provide positive instruction feedback to LLM, the token is better than the last one.
    \ENDIF
    \STATE $A^P = A_t^P{'}$.
\ENDFOR
\end{algorithmic}
\end{algorithm}

\section{Experiments}
\label{section:experiments}

Our experimental study is conducted within the BabyAI benchmark~\citep{babyaiiclr19}, utilizing the Lamorel framework~\citep{carta2023grounding} to support our investigation. Within this framework, we integrate RL instruction feedback into LLMs, establishing a feedback loop to enhance LLM performance. Specifically, we focus on the GoToRedBallNoDists-v0 planning task, conducting experiments under two conditions: one with 40 iteration steps and another with 2100 iteration steps. We perform a comparative analysis between our proposed method and the baseline for evaluation. Our approach incorporates bidirectional interaction, where RL models provide feedback to LLMs, and LLMs supply information to RL models. In contrast, the state-of-the-art baseline, represented by the original Lamorel method, lacks this feedback mechanism from RL models to LLMs.

\begin{figure}[htbp!]
 \centering
 \subcaptionbox{}
 {
\includegraphics[width=0.45\linewidth]{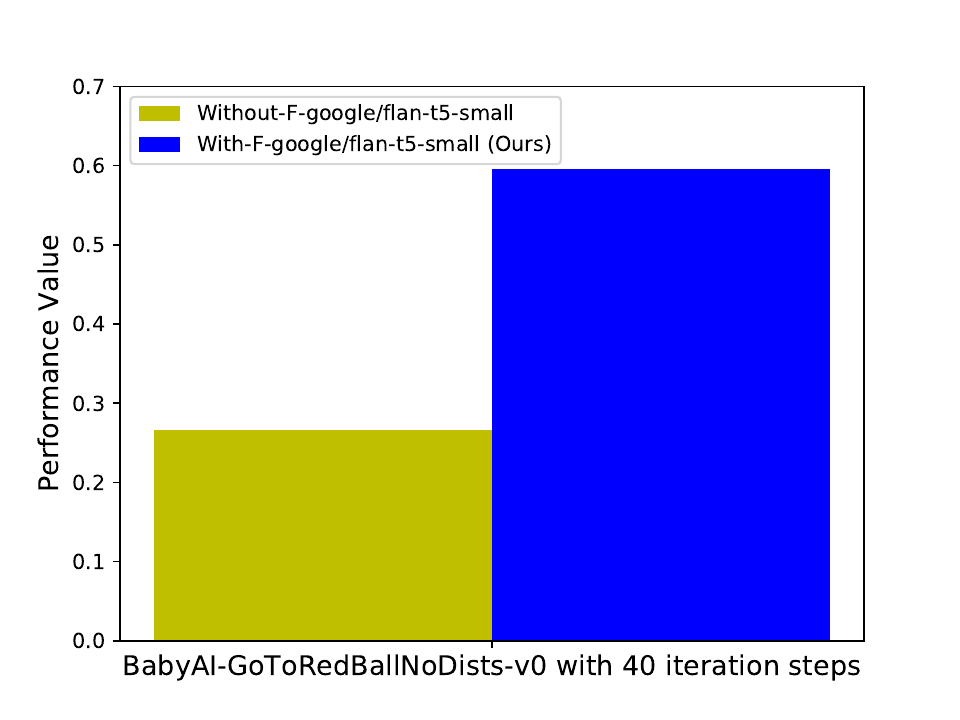}
}
 \subcaptionbox{}
 {
\includegraphics[width=0.45\linewidth]{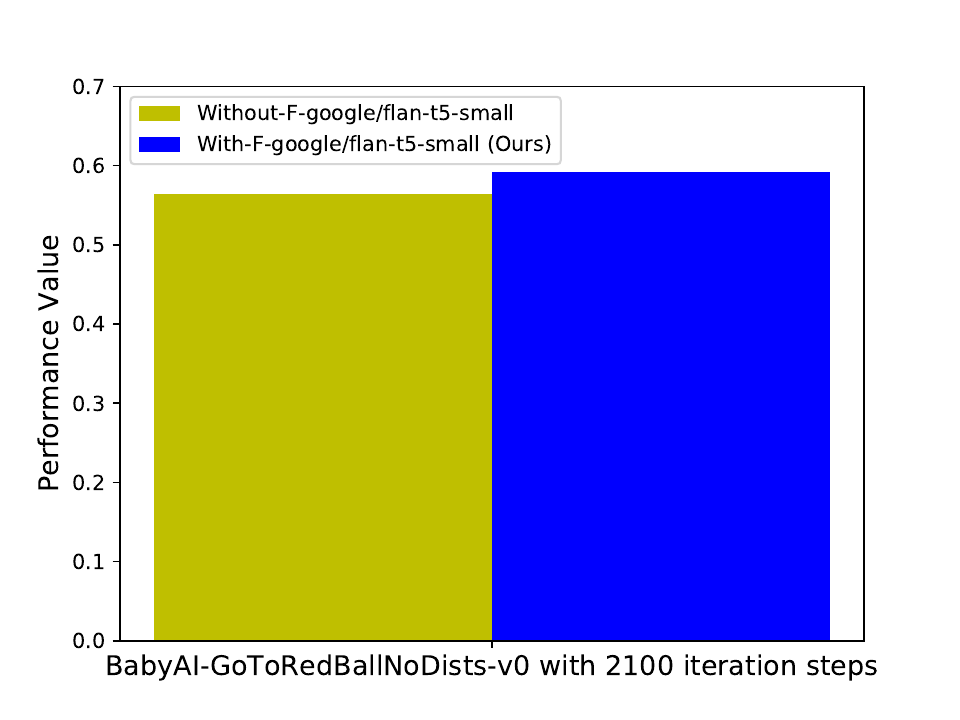}
}
    \vspace{-0pt}
 	\caption{\normalsize 
  Experiments on BabyAI tasks~\cite{babyaiiclr19} with 40 (a) and 2100 (b) iteration steps.
 	} 
  \label{fig:experiment-rl-llm-babyai-results}
 \end{figure}

It is noteworthy that for our experiments, we employ the "google/flan-t5-small" model~\citep{chung2022scaling} as the LLM, characterized by a parameter count of 80 million. The experimental results are presented in Figure \ref{fig:experiment-rl-llm-babyai-results}. These findings clearly illustrate the superior performance of our method, as quantified by the performance value metric (where higher values indicate better performance). Furthermore, our method demonstrates notably expedited convergence (one-shot/few-shot learning) when compared to the Lamorel baseline. This empirical evidence highlights the effectiveness of our approach in harnessing bi-directional feedback between RL models and LLMs to improve performance in the context of the BabyAI benchmark.

\section{Conclusion}
\label{section:conclusion}

In this study, we developed a teacher-student learning framework for unlocking LLMs' powerful capabilities by leveraging an RL model with bi-directional feedback mechanisms in a cooperative game setting. To empirically assess the effectiveness of our method, we conducted experiments using the BabyAI benchmark as an assessment platform. The results of these experiments demonstrate the superior performance of our approach in comparison to the state-of-the-art baseline, highlighting its potential for substantially enhancing learning outcomes. Notably, our approach holds promise for fostering safe and robust learning systems~\citep{gu2022review}, particularly in environments characterized by imperfect information. Furthermore, we hope that our findings inspire novel research directions at the intersection of LLMs and RL. As part of future work, we plan to extend our method to more challenging tasks and assess its effectiveness in complex real-world applications.

\normalem






\bibliography{main}
\bibliographystyle{plain}

\end{document}